%% file: root.tex
\documentclass[letterpaper, 10 pt, journal, twoside]{IEEEtran} 

\IEEEoverridecommandlockouts                              

\usepackage[normalem]{ulem}
\usepackage[table,usenames,dvipsnames]{xcolor}
\usepackage{extarrows}
\let\labelindent\relax
\usepackage[inline]{enumitem}
\usepackage{cite}

\usepackage{amsmath,amssymb,amsfonts,dsfont} 
\usepackage{mathtools, bm}
\usepackage{algorithm,algorithmicx,listings}
\usepackage[noend]{algpseudocode}

\newcommand{\RR}{\mathbb{R}}

\usepackage{graphicx}
\usepackage{tabularx}
\usepackage[font={small}]{caption}   
\usepackage{subcaption}
\usepackage[breaklinks=true, colorlinks, bookmarks=true, citecolor=Black, urlcolor=Violet,linkcolor=Black]{hyperref}

\title{
Robust Stereo Visual Inertial Odometry for Fast Autonomous Flight
}

\author{Ke Sun, Kartik Mohta, Bernd Pfrommer, Michael Watterson, Sikang Liu, \\
Yash Mulgaonkar, Camillo J. Taylor, and Vijay Kumar*
\thanks{Manuscript received: September, 10, 2017; Revised December, 1, 2017; Accepted December, 27, 2017. This paper was recommended for publication by Editor Cyrill Stachniss upon evaluation of the Associate Editor and Reviewers' comments. This work was supported in part by DARPA grants HR001151626 and HR0011516850, ARL grant W911NF-08-2-0004, and a NASA Space Technology Research Fellowship awarded to Michael Watterson.}
\thanks{*The authors are with GRASP Lab, University of Pennsylvania, Philadelphia, PA 19104, USA, {\tt\small\{sunke, kmohta, pfrommer, wami, sikang, yashm, cjtaylor, kumar\}@seas.upenn.edu}.}
\thanks{Digital Object Identifier (DOI): see top of this page.}
}

\begin{document}

\maketitle
\markboth{IEEE Robotics and Automation Letters. Preprint Version. Accepted December, 2017}
{Sun \MakeLowercase{\textit{et al.}}: Robust Stereo Visual Inertial Odometry for Fast Autonomous Flight}

\begin{abstract}
In recent years, vision-aided inertial odometry for state estimation has matured significantly. However, we still encounter challenges in terms of improving the computational efficiency and robustness of the underlying algorithms for applications in autonomous flight with micro aerial vehicles in which it is difficult to use high quality sensors and powerful processors because of constraints on size and weight. In this paper, we present a filter-based stereo visual inertial odometry that uses the Multi-State Constraint Kalman Filter (MSCKF)~\cite{mourikis2007multi}. Previous work on stereo visual inertial odometry has resulted in solutions that are computationally expensive. We demonstrate that our Stereo Multi-State Constraint Kalman Filter (S-MSCKF) is comparable to state-of-art monocular solutions in terms of computational cost, while providing significantly greater robustness. We evaluate our S-MSCKF algorithm and compare it with state-of-art methods including OKVIS, ROVIO, and VINS-MONO on both the EuRoC dataset, and our own experimental datasets demonstrating fast autonomous flight with maximum speed of $17.5$m/s in indoor and outdoor environments. Our implementation of the S-MSCKF is available at \url{https://github.com/KumarRobotics/msckf_vio}.
\end{abstract}

\input{tex/Introduction.tex}
\input{tex/RelatedWork.tex}

\input{tex/FilterDescription.tex}

\input{tex/Experiments.tex}
\input{tex/Conclusion.tex}


\section*{APPENDIX}
\appendices
\input{tex/Appendices}


\bibliographystyle{IEEEtran}
\bibliography{ref}

\end{document}

%% file: tex/Introduction.tex
\section{Introduction}
\label{sec: introduction}
\IEEEPARstart{A}{ccurate} and robust state estimation is of crucial importance for robot autonomy and in particular for micro aerial vehicles (MAVs), where correct pose estimation is essential for stabilizing the robot in the air.

The solution of combining visual information from cameras and measurements from an Inertial Measurement Unit (IMU), usually referred to as Visual Inertial Odometry (VIO), is popular because it can perform well in GPS-denied environments and, compared to lidar based approaches, requires only a small and  lightweight sensor package, making it the preferred technique for MAV platforms.

In scenarios such as search and rescue or first response, MAVs have to operate in a wide range of environments that pose challenges to VIO algorithms such as drastically varying lighting conditions, uneven illumination, low texture scenes, and abrupt changes in attitude due to wind gusts or aggressive maneuvering. Thus the VIO not only has to be accurate, it must also be robust.

\begin{figure}[tp]
\centering
\includegraphics[width=0.4\textwidth]{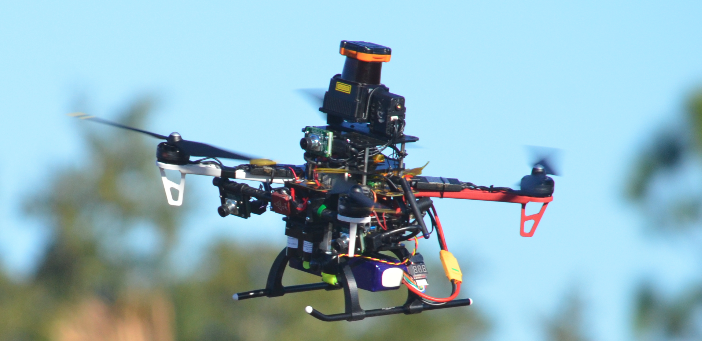}
\caption{The 3kg {\sc Falcon} robot with an onboard Intel NUC5i7RYH computer,  synchronized stereo cameras and IMU, a laser scanner, and a downward facing lidar. Note that only the stereo cameras and the IMU are used for state estimation.}
\label{fig: fla robot}
\end{figure}

In order to achieve full autonomy, all software components, from low-level sensor drivers to high-level planning algorithms, have to run onboard in real-time on a computer with similar computational power to a laptop because of the limited payload of a MAV such as the one shown in Figure~\ref{fig: fla robot}. The requirement to share onboard resources with other components puts additional pressure on the VIO algorithm to be as efficient as possible and importantly, to not produce excessive intermittent spikes in CPU consumption.

In this paper, we propose a filter-based stereo VIO solution to address such challenges, mostly because they are generally more computationally efficient than competing optimization-based methods. Among the filtering approaches, we choose as a starting point the state-of-art MSCKF~\cite{mourikis2007multi,li2013high,li2013optimization} algorithm for its accuracy and consistency. A stereo configuration is preferred over the recently more popular monocular solutions because of its robustness to different environments and motion. Contradicting the widely held belief that stereo vision-based estimation incurs much higher compute cost than monocular approaches, we demonstrate that the proposed stereo VIO is able to achieve similar or even higher efficiency than state-of-art monocular solutions. Our main contributions are:
\begin{itemize}
\item To the best of our knowledge, the presented work is the first open-source filter-based stereo VIO that can run onboard on a laptop-class computer without GPU acceleration. 
\item We provide detailed experimental comparisons between the proposed S-MSCKF and state-of-art open-source VIO solutions including OKVIS~\cite{leutenegger2015keyframe}, ROVIO~\cite{bloesch2015robust}, and VINS-MONO~\cite{vins-mono} in accuracy, efficiency and robustness. The comparison is performed on both the EuRoC~\cite{burri2016euroc} dataset and the fast flight dataset using {\sc Falcon} shown in Figure~\ref{fig: fla robot}. 
\item The fast flight dataset is publicly available at \url{https://github.com/KumarRobotics/msckf_vio/wiki}.
\end{itemize}

%
%
The rest of the paper is organized as follows: Section~\ref{sec: related work} discusses the related work. Section~\ref{sec: filter description} presents the mathematical details of the proposed algorithm. In Section~\ref{sec: experiments}, we compare our work with different state-of-the-art works in VIO on different datasets and demonstrate the performance of the proposed S-MSCKF with a fully autonomous flight through unstructured and unknown environments. Finally, Section~\ref{sec: conclusion} concludes the paper.

%% file: tex/RelatedWork.tex
\section{Related Work}
\label{sec: related work}
There is extensive literature on visual odometry, ranging from pure vision-based methods~\cite{forster2014svo, Geiger2011IV, mur2017orb, engel2017direct, engel2014lsd}, loosely-coupled VIO solutions~\cite{weiss2012real, shen2014multi}, to the recently more popular tightly-coupled VIO solutions~\cite{leutenegger2015keyframe, yang2017monocular, usenko2016direct, forster2015imu, mourikis2007multi, wu2015square, bloesch2015robust, kelly2011visual, tsotsos2015robust} which will be the major focus of this paper. 

Existing tightly-coupled VIO solutions can, in general, be classified into optimization-based (e.g. \cite{leutenegger2015keyframe, yang2017monocular, usenko2016direct}) and filter-based approaches (e.g. \cite{mourikis2007multi, tsotsos2015robust, bloesch2015robust}). Optimization-based methods obtain the optimal estimate by jointly minimizing the residual using the measurements from both IMU data and images. Most of these systems, such as~\cite{yang2017monocular}, use sparse features obtained from images as measurements. Methods like this are also called indirect methods. Usenko et al.~\cite{usenko2016direct} and Forster et al.~\cite{forster2015imu} propose a direct method which minimizes photometric error directly in order to exploit more information from the images. The literature shows that optimization-based approaches are able to achieve high accuracy. However, such methods require significant computational resources because of the iterative optimization process, although recent efficient solvers (e.g.~\cite{kaess2012isam2, ceres-solver}) can be run in real time online.

 In contrast, filter-based approaches, which generally use the Extended Kalman Filter (EKF)~\cite{mourikis2007multi}, or the Uncented Kalman Filter~\cite{kelly2011visual}, are much more efficient while achieving accuracy comparable to optimization-based approaches. Huang et al.~\cite{huang2010observability}, Li et al.~\cite{li2013high}, and Hesch et al.~\cite{hesch2014consistency}  also propose the First Estimate Jacobian (FEJ) and the Observability Constraint (OC) to improve consistency of VIO in the filter framework, which in turn improves the estimation accuracy. In  more recent work~\cite{bloesch2015robust, xing2017photometric}, the direct method is used in the filter-based VIO framework to further improve accuracy and robustness.

Only a few VIO solutions are designed for stereo or multi-camera system~\cite{lupton2012visual, leutenegger2015keyframe, usenko2016direct, paulcomparative} compared to the vast amount of work on monocular systems. This can be attributed in part to costs associated with processing additional images and matching features.
In \cite{lupton2012visual}, datasets are collected using stereo visual inertial configuration with cameras running at $6.25$hz, which are then processed offline. The stereo VIO in~\cite{lupton2012visual} is more a proof of concept of IMU pre-integration and does not lend itself to practical implementation. Leutenegger et al.~\cite{leutenegger2015keyframe} propose a more complete optimization framework for multi-camera VIO that is able to run in real time. Usenko~\cite{usenko2016direct} introduces direct methods into stereo VIO in order to further improve accuracy. All three solutions are optimization-based approaches requiring powerful CPUs to operate in real time. More recently, Paul et al.~\cite{paulcomparative} proposed a filter-based stereo VIO based on the square root inverse filter~\cite{wu2015square}, which demonstrates the possibility of operating a stereo VIO online efficiently, even on a mobile device. Since the implementations of~\cite{usenko2016direct} and~\cite{wu2015square} are not open-sourced, they are not used for comparison in this paper.

%% file: tex/FilterDescription.tex
\section{Filter Description}
\label{sec: filter description}
In the description of the filter setup, we follow the convention in~\cite{mourikis2007multi}. The IMU state is defined as,
\begin{equation*}
\mathbf{x}_{I} = 
\left(
{}^I_G \mathbf{q}^\top \quad 
\mathbf{b}_g^\top \quad 
{}^G\mathbf{v}^\top_I \quad 
\mathbf{b}_a^\top \quad
{}^G\mathbf{p}^\top_I \quad
{}^I_C \mathbf{q}^\top \quad
{}^I\mathbf{p}^\top_C
\right)^\top
\end{equation*}
where the quaternion ${}^I_G \mathbf{q}$ represents the rotation from the inertial frame to the body frame. In our configuration, the body frame is set to be the IMU frame. The vectors ${}^G\mathbf{v}_I \in \RR^3$ and ${}^G\mathbf{p}_I \in \RR^3$ represent the velocity and position of the body frame in the inertial frame. The vectors $\mathbf{b}_g \in \RR^3$ and $\mathbf{b}_a \in \RR^3$ are the biases of the measured angular velocity and linear acceleration from the IMU. Finally the quaternion, ${}^I_C \mathbf{q}$ and ${}^I\mathbf{p}_C \in \RR^3$ represent the relative transformation between the camera frame and the body frame. Without loss of generality, the left camera frame is used assuming the extrinsic parameters relating the left and right cameras are known. Using the true IMU state would cause singularities in the resulting covariance matrices because of the additional unit constraint on the quaternions in the state vector. Instead, the error IMU state, defined as,
\begin{equation*}
\tilde{\mathbf{x}}_{I} = 
\left(
{}^I_G \tilde{\bm{\theta}}^\top \quad 
\tilde{\mathbf{b}}_g^\top \quad 
{}^G\tilde{\mathbf{v}}^\top_I \quad 
\tilde{\mathbf{b}}_a^\top \quad
{}^G\tilde{\mathbf{p}}^\top_I \quad
{}^I_C \tilde{\bm{\theta}}^\top \quad
{}^I\tilde{\mathbf{p}}^\top_C
\right)^\top
\end{equation*}
is used with standard additive error used for position, velocity, and biases (e.g. ${}^G\tilde{\mathbf{p}}_I = {}^G\mathbf{p}_I-{}^G\hat{\mathbf{p}}_I$). For the quaternions, the error quaternion $\delta\mathbf{q} = \mathbf{q}\otimes\hat{\mathbf{q}}^{-1}$ is related to the error state as,
\begin{equation*}
\delta\mathbf{q} \approx
\left(
\frac{1}{2} {}^G_I\tilde{\bm{\theta}}^\top \quad 1
\right)^\top
\end{equation*}
where ${}^G_I\tilde{\bm{\theta}} \in \RR^3$ represents a
small angle rotation. With such a representation, the dimension of orientation error is reduced to $3$ enabling proper presentation of its uncertainty. Ultimately $N$ camera states are considered together in the state vector, so the entire error state vector would be,
\begin{equation*}
\tilde{\mathbf{x}} = 
\left(
\tilde{\mathbf{x}}_I^\top \quad
\tilde{\mathbf{x}}_{C_1}^\top \quad
\cdots \quad 
\tilde{\mathbf{x}}_{C_N}^\top
\right)^\top
\end{equation*}
where each camera error state is defined as,
\begin{equation*}
\tilde{\mathbf{x}}_{C_i} = 
\left(
{}^{C_i}_G\tilde{\bm{\theta}}^\top \quad
{}^G\tilde{\mathbf{p}}_{C_i}^\top
\right)^\top
\end{equation*}
In order to maintain bounded computational complexity, some camera states have to be marginalized once the number of camera states reaches a preset limit. Discussions of how to choose camera states to marginalize can be found in Section \ref{subsec: filter update mechanism}. 
\input{tex/ProcessModel.tex}
\input{tex/MeasurementModel.tex}

\input{tex/ObservabilityConstraint.tex}
\input{tex/FilterUpdateMechanism.tex}
\input{tex/ImageProcessingFrontend.tex}

%% file: tex/ProcessModel.tex
\subsection{Process Model}
\label{subsec: process model}
The continuous dynamics of the estimated IMU state is,
\begin{equation}
\label{eq: estimated state dynamics}
\begin{gathered}
{}^I_G\dot{\hat{\mathbf{q}}} = \frac{1}{2}\Omega(\hat{\bm{\omega}}) {}^I_G\hat{\mathbf{q}}, \quad
\dot{\hat{\mathbf{b}}}_g = \mathbf{0}_{3\times 1}, \\
{}^G\dot{\hat{\mathbf{v}}} = C\left({}^I_G\hat{\mathbf{q}}\right)^\top \hat{\mathbf{a}} + {}^G\mathbf{g}, \\
\dot{\hat{b}}_a = \mathbf{0}_{3\times 1}, \quad
{}^G\dot{\hat{\mathbf{p}}}_I = {}^G\hat{\mathbf{v}}, \\
{}^I_C\dot{\hat{\mathbf{q}}} = \mathbf{0}_{3\times 1}, \quad
{}^I\dot{\hat{\mathbf{p}}}_C = \mathbf{0}_{3\times 1}
\end{gathered}
\end{equation}
where $\hat{\bm{\omega}} \in \RR^3$ and $\hat{\mathbf{a}} \in \RR^3$ are the IMU measurements for angular velocity and acceleration respectively with biases removed, i.e,
\begin{equation*}
\hat{\bm{\omega}} = \bm{\omega}_m - \hat{\mathbf{b}}_g, \quad
\hat{\mathbf{a}} = \mathbf{a}_m - \hat{\mathbf{b}}_a
\end{equation*}
Meanwhile,
\begin{equation*}
\Omega\left(\hat{\bm{\omega}}\right) = 
\begin{pmatrix}
-[\hat{\bm{\omega}}_\times] & \bm{\omega} \\
-\bm{\omega}^\top & 0
\end{pmatrix}
\end{equation*}
where $[\hat{\bm{\omega}}_\times]$ is the skew symmetric matrix of $\hat{\bm{\omega}}$. $C(\cdot)$ in Eq.~\eqref{eq: estimated state dynamics} is the function converting quaternion to the corresponding rotation matrix. Based on Eq.~\eqref{eq: estimated state dynamics}, the linearized continuous dynamics for the error IMU state follows,
\begin{equation}
\label{eq: error state dynamics}
\dot{\tilde{\mathbf{x}}}_I = 
\mathbf{F} \tilde{\mathbf{x}}_I + 
\mathbf{G} \mathbf{n}_I
\end{equation}
where $\mathbf{n}_I^\top = \left(\mathbf{n}_g^\top \; \mathbf{n}_{wg}^\top \; \mathbf{n}_a^\top \; \mathbf{n}_{wa}^\top\right)^\top$. The vectors $\mathbf{n}_g$ and $\mathbf{n}_a$ represent the Gaussian noise of the gyroscope and accelerometer measurement, while $\mathbf{n}_{wg}$ and $\mathbf{n}_{wa}$ are the random walk rate of the gyroscope and accelerometer measurement biases. $\mathbf{F}$ and $\mathbf{G}$ are shown in Appendix~\ref{sec: error state dynamics}.

To deal with discrete time measurement from the IMU, we apply a $4^{th}$ order Runge-Kutta numerical integration of Eq.~\eqref{eq: estimated state dynamics} to propagate the estimated IMU state. To propagate the uncertainty of the state, the discrete time state transition matrix of Eq.~\eqref{eq: error state dynamics} and discrete time noise covairance matrix need to be computed first, 
\begin{equation*}
\begin{gathered}
\bm{\Phi}_k = \bm{\Phi}(t_{k+1}, t_k) = 
\exp\left(\int_{t_k}^{t_{k+1}} \mathbf{F}(\tau)d\tau\right) \\
\mathbf{Q}_k = \int_{t_k}^{t_{k+1}} \bm{\Phi}(t_{k+1},\tau)\mathbf{G}\mathbf{Q}\mathbf{G}\bm{\Phi}(t_{k+1},\tau)^\top d\tau
\end{gathered}
\end{equation*}
where $\mathbf{Q} = \mathbb{E}\left[\mathbf{n}_I^{}\mathbf{n}_I^\top\right]$ is the continuous time noise covariance matrix of the system. Then the propagated covariance of the IMU state is,
\begin{equation*}
\mathbf{P}_{II_{k+1|k}} = \bm{\Phi}_k\mathbf{P}_{II_{k|k}}\bm{\Phi}_k^\top + \mathbf{Q}_k
\end{equation*}
By partioning the covariance of the whole state as,
\begin{equation*}
\mathbf{P}_{k|k} = 
\begin{pmatrix}
\mathbf{P}_{II_{k|k}} & \mathbf{P}_{IC_{k|k}} \\
\mathbf{P}_{IC_{k|k}}^\top & \mathbf{P}_{CC_{k|k}}
\end{pmatrix}
\end{equation*}
the full uncertainty propagation can be represented as,
\begin{equation*}
\mathbf{P}_{k+1|k} = 
\begin{pmatrix}
\mathbf{P}_{II_{k+1|k}} & \bm{\Phi}_k \mathbf{P}_{IC_{k|k}} \\
\mathbf{P}_{IC_{k|k}}^\top \bm{\Phi}_k^\top & \mathbf{P}_{CC_{k|k}}
\end{pmatrix}
\end{equation*}
When new images are received, the state should be augmented with the new camera state. The pose of the new camera state can be computed from the latest IMU state as,
\begin{equation*}
{}^C_G\hat{\mathbf{q}} = {}^C_I\hat{\mathbf{q}} \otimes {}^I_G\hat{\mathbf{q}}, \quad
{}^G\hat{\mathbf{p}}_C = {}^G\hat{\mathbf{p}}_C + C\left({}^I_G\hat{\mathbf{q}}\right)^\top {}^I\hat{\mathbf{p}}_C
\end{equation*}
And the augmented covariance matrix is,
\begin{equation}
\label{eq: state covariance augmentation}
\mathbf{P}_{k|k} = 
\begin{pmatrix}
\mathbf{I}_{21+6N} \\ \mathbf{J}
\end{pmatrix}
\mathbf{P}_{k|k}
\begin{pmatrix}
\mathbf{I}_{21+6N} \\ \mathbf{J}
\end{pmatrix}^\top
\end{equation}
where $\mathbf{J}$ is shown in Appendix~\ref{sec: state augmentation jacobian}.

%% file: tex/MeasurementModel.tex
\subsection{Measurement Model}
\label{subsec: measurement model}
Consider the case of a single feature $f_j$ observed by the stereo cameras with pose $\left({}^{C_i}_G\mathbf{q}, {}^G\mathbf{p}_{C_i}\right)$. Note that the stereo cameras have different poses, represented as $\left({}^{C_{i,1}}_G\mathbf{q}, {}^G\mathbf{p}_{C_{i,1}}\right)$ and $\left({}^{C_{i,2}}_G\mathbf{q}, {}^G\mathbf{p}_{C_{i,2}}\right)$ for left and right cameras respectively, at the same time instance. Although the state vector only contains the pose of the left camera, the pose of the right camera can be easily obtained using the extrinsic parameters from the calibration. The stereo measurement, $\mathbf{z}_i^j$, is represented as,
\begin{equation}
\mathbf{z}_i^j = 
\begin{pmatrix}
u_{i, 1}^j \\ v_{i, 1}^j \\ 
u_{i, 2}^j \\ v_{i, 2}^j
\end{pmatrix} 
= 
\begin{pmatrix}
\frac{1}{{}^{C_{i, 1}}Z_j} & \mathbf{0}_{2\times 2} \\
\mathbf{0}_{2\times 2} & \frac{1}{{}^{C_{i, 2}}Z_j}
\end{pmatrix}
\begin{pmatrix}
{}^{C_{i, 1}}X_j \\ {}^{C_{i, 1}}Y_j \\
{}^{C_{i, 2}}X_j \\ {}^{C_{i, 2}}Y_j
\end{pmatrix}
\label{eq: stereo measurement}
\end{equation}
Note that the dimension of $\mathbf{z}_i^j$ can be reduced to $\mathbb{R}^3$ assuming the stereo images are properly rectified. However, by representing $\mathbf{z}_i^j$ in $\mathbb{R}^4$, it is no longer required that the observations of the same feature on the stereo images are on the same image plane, which removes the necessity for stereo rectification. In Eq.~\eqref{eq: stereo measurement}, $\left({}^{C_{i, k}}X_j\  {}^{C_{i, k}}Y_j\  {}^{C_{i, k}}Z_j \right)^\top$, $k\in\{1, 2\}$, are the positions of the feature, $f_j$, in the left and right camera frame, $C_{i, 1}$ and $C_{i, 2}$, which are related to the camera pose by,
\begin{equation*}
\begin{gathered}
{}^{C_{i, 1}}\mathbf{p}_j = 
\begin{pmatrix}
{}^{C_{i, 1}}X_j \\ {}^{C_{i, 1}}Y_j \\ {}^{C_{i, 1}}Z_j
\end{pmatrix} = 
C\left({}^{C_{i, 1}}_G\mathbf{q}\right)
\left({}^G\mathbf{p}_j-{}^G\mathbf{p}_{C_{i, 1}}\right) \\
\begin{aligned}
{}^{C_{i, 2}}\mathbf{p}_j &= 
\begin{pmatrix}
{}^{C_{i, 2}}X_j \\ {}^{C_{i, 2}}Y_j \\ {}^{C_{i, 2}}Z_j
\end{pmatrix} = 
C\left({}^{C_{i, 2}}_G\mathbf{q}\right)
\left({}^G\mathbf{p}_j-{}^G\mathbf{p}_{C_{i, 2}}\right) \\
&= C\left({}^{C_{i, 2}}_{C_{i, 1}}\mathbf{q}\right)
\left({}^{C_{i, 1}}\mathbf{p}_j - 
{}^{C_{i, 1}}\mathbf{p}_{C_{i, 2}}\right)
\end{aligned}
\end{gathered}
\end{equation*} 
The position of the feature in the world frame, ${}^G\mathbf{p}_j$, is 
computed using the least square method given in~\cite{mourikis2007multi} based on the current estimated camera poses. Linearizing the measurement model at the current estimate, the residual of the measurement can be approximated as,
\begin{equation}
\label{eq: error measurement model}
\mathbf{r}^j_i = 
\mathbf{z}_i^j - \hat{\mathbf{z}}_i^j = 
\mathbf{H}_{C_i}^j\tilde{\mathbf{x}}_{C_i} + 
\mathbf{H}_{f_i}^j{}^G\tilde{\mathbf{p}}_{j} + 
\mathbf{n}_i^j
\end{equation}
where $\mathbf{n}_i^j$ is the noise of the measurement. The measurement Jacobian $\mathbf{H}_{C_i}^j$ and $\mathbf{H}_{f_i}^j$ are shown in Appendix~\ref{sec: measurement jacobian}.

By stacking multiple observations of the same feature $f_j$, we have,
\begin{equation*}
\mathbf{r}^j = 
\mathbf{H}_{\mathbf{x}}^j\tilde{\mathbf{x}} + 
\mathbf{H}_f^j {}^G\tilde{\mathbf{p}}_j + 
\mathbf{n}^j
\end{equation*}
As pointed out in~\cite{mourikis2007multi}, since ${}^G\mathbf{p}_j$ is computed using the camera poses, the uncertainty of ${}^G\mathbf{p}_j$ is, therefore, correlated with the camera poses in the state. In order to ensure the uncertainty of ${}^G\mathbf{p}_j$ does not affect the residual, the residual in Eq.~\eqref{eq: error measurement model} is projected to the null space, $\mathbf{V}$, of $\mathbf{H}_f^j$ , i.e.
\begin{equation}
\label{eq: null space error measurement model}
\mathbf{r}^j_o 
= \mathbf{V}^\top \mathbf{r}^j
= \mathbf{V}^\top \mathbf{H}_{\mathbf{x}}^j\tilde{\mathbf{x}} +
\mathbf{V}^\top \mathbf{n}^j
= \mathbf{H}_{\mathbf{x}, o}^j\tilde{\mathbf{x}} + 
\mathbf{n}^j_o
\end{equation}
Based on Eq.~\eqref{eq: null space error measurement model}, the update step of the EKF can be carried out in a standard way.

%% file: tex/ObservabilityConstraint.tex
\subsection{Observability Constraint}
\label{subsec: observability constraint}
As has been shown in~\cite{huang2010observability, li2013high}, the EKF-based VIO for 6-DOF motion estimation has four unobservable directions corresponding to global position and rotation along the gravity axis, i.e. yaw angle. A naive implementation of EKF VIO will gain spurious information on yaw. This is due to the fact that the linearizing point of the process and measurement step are different at the same time instant.

There are different methods for maintaining the consistency of the filter, including the First Estimate Jacobian EKF (FEJ-EKF)~\cite{huang2010observability}, the Observability Constrained EKF (OC-EKF)~\cite{hesch2012observability}, and Robocentric Mapping Filter~\cite{castellanos2007robocentric}. In our implementation, OC-EKF is applied for two reasons as discussed in~\cite{huang2010observability},
\begin{enumerate*}[label=(\roman*)]
\item unlike FEJ-EKF, OC-EKF does not heavily depend on an accurate initial estimation,
\item comparing to Robocentric Mapping Filter, camera poses in the state vector can be represented with respect to the inertial frame instead of the latest IMU frame so that the uncertainty of the existing camera states in the state vector is not affected by the uncertainty of the latest IMU state during the propagation step.
\end{enumerate*}

%% file: tex/FilterUpdateMechanism.tex
\subsection{Filter Update Mechanism}
\label{subsec: filter update mechanism}
Two types of delayed measurement updates are described in~\cite{mourikis2007multi}. The measurement update step is executed when either the algorithm loses a feature or the number of camera poses in the state reaches the limit. In our implementation, we inherit the same update mechanism with modifications for real-time considerations. As suggested in~\cite{mourikis2007multi}, one third of the camera states are marginalized once the buffer is full, which can cause sudden jumps in computational load in real-time implementations. It is desired that one camera state is marginalized at each time step in order to average out the computation. However, removing the observation of a feature at one camera state is not practical in the MSCKF framework since the observation contains no information about the relative transformation between the camera states. Mathematically, this is due to the fact that the null space of $\mathbf{H}_{f_i}^j$ in Eq.~\eqref{eq: error measurement model} is a subspace of the null space of $\mathbf{H}_{C_i}^j$ (see Appendix~\ref{sec: nullify measurement jacobian}), which results in a trivial measurement model based on Eq.~\eqref{eq: error measurement model}. 

In our implementation, two camera states are removed every other update step. All feature observations obtained at the two camera states are used for measurement update. Note that because of the reason mentioned above, the two stereo measurements of the two camera states are only useful if they are of the same feature. It is understood that such frequent removal of camera states can cause some valid observations to be ignored. In practice, we found that the estimation performance is barely affected although fewer observations are used. To select the two camera states to be removed, we apply a keyframe selection strategy similar to the two-way marginalization method proposed in~\cite{shen2014initialization}.Based on the relative motion between the second latest camera state and its previous one, either the second latest or the oldest camera state is selected for removal. The selection procedure is executed twice to find two camera states to remove. Note that the latest camera state is always kept since it has the measurements for the newly detected features.

%% file: tex/ImageProcessingFrontend.tex
\subsection{Image Processing Frontend}
\label{subsec: image processing frontend}
In our implementation, the FAST~\cite{trajkovic1998fast} feature detector is employed for its efficiency. Existing features are tracked temporally using the KLT optical flow algorithm~\cite{lucas1981iterative}. It is shown in~\cite{paulcomparative} that descriptor-based methods for temporal feature tracking are better than KLT-based methods in accuracy. In our experiments, we find that descriptor-based methods require much more CPU resource with small gain in accuracy, making such methods less favorable in our application. Note that we also use the KLT optical flow algorithm for stereo feature matching, which further saves computation compared to descriptor-based methods. Empirically, corner features with depths greater than $1$m can be reliably matched across the stereo images using KLT tracking with a $20$cm baseline stereo configuration. Two types of outlier rejection procedure are implemented in the image processing frontend. A 2-point RANSAC is applied to remove outliers in temporal tracking. In addition, a circular matching similar to~\cite{kitt2010visual} is performed between the previous and current stereo image pairs to further remove outliers generated in the feature tracking and stereo matching steps.

%% file: tex/Experiments.tex
\section{Experiments}
\label{sec: experiments}
In order to evaluate the proposed method, three different kinds of experiments were performed. First, the proposed method was compared with  state-of-art visual inertial odometry algorithms including OKVIS~\cite{leutenegger2015keyframe}, ROVIO~\cite{bloesch2015robust}, and VINS-MONO~\cite{vins-mono} on the EuRoC dataset~\cite{burri2016euroc}. Second, we demonstrate the robustness of the proposed algorithm on high speed flights reaching maximum speeds of $17.5$m/s on a runway environment. In both of the experiments, the loop closure functionality of VINS-MONO is disabled in order to just compare the odometry of different approaches. Note that although all of the algorithms used in the comparison are capable of estimating extrinsic parameters between the IMU and camera frames online, the offline calibration parameters are provided in the experiments for optimal performance. Finally, we show a representative application of the proposed S-MSCKF in an experiment that combines estimation, with control and planning for autonomous flight in an unstructured and unknown environment which includes a warehouse, a wooded area and a runway.
\input{tex/EurocDataset.tex}
\input{tex/FastFlightDataset.tex}

\input{tex/FlaFieldTest.tex}

%% file: tex/EurocDataset.tex
\subsection{EuRoC Dataset}
\label{subsec: euroc dataset}
The EuRoC datasets were collected with a VI sensor~\cite{nikolic2014synchronized} on a MAV, which includes synchronized $20$Hz stereo images and $200$Hz IMU messages. The aggressive rotation and significant lighting change make the dataset challenging for vision-based state estimation. 
\begin{figure}[htp]
\centering
\begin{subfigure}[b]{0.4\textwidth}
\includegraphics[width=\textwidth]{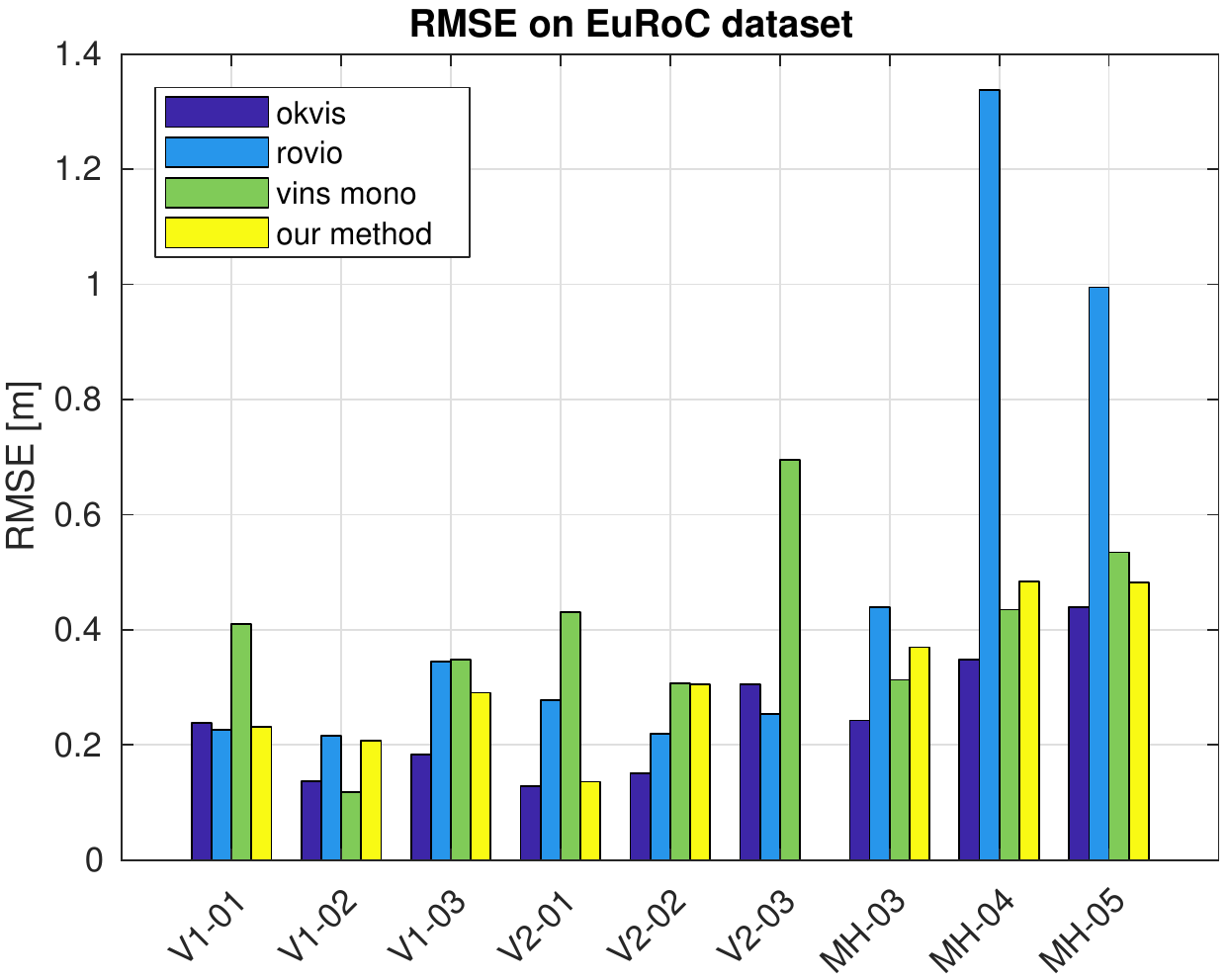}
\caption{•}
\label{fig: euroc accuracy benchmark}
\end{subfigure}
\begin{subfigure}[b]{0.4\textwidth}
\includegraphics[width=\textwidth]{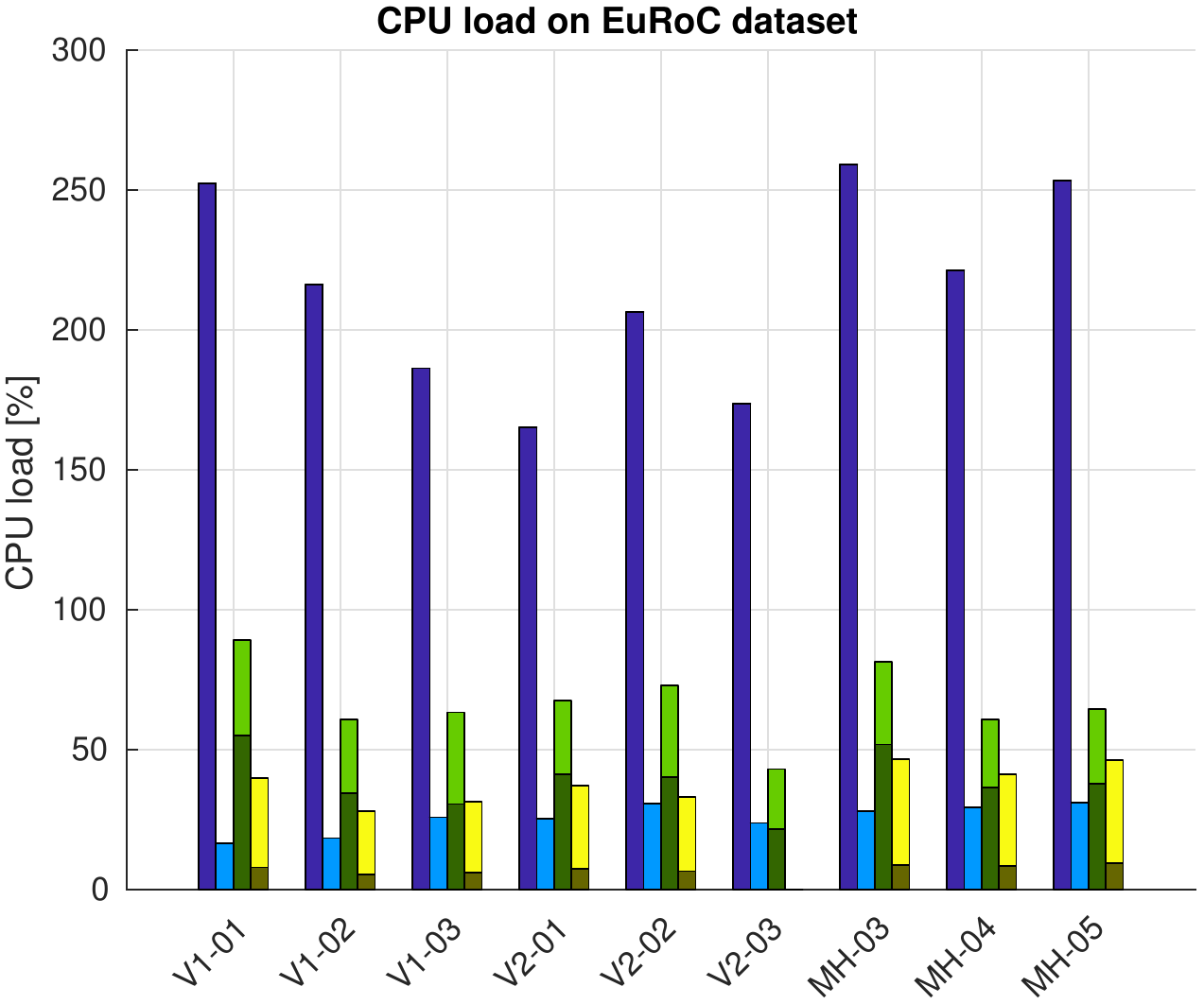}
\caption{•}
\label{fig: euroc cpu benchmark}
\end{subfigure}
\caption{(a) Root Mean Square Error (RMSE) and (b) average CPU load of OKVIS, ROVIO, VINS-MONO, and the proposed method on the EuRoC dataset. The parameters used for each method are the same as the values given in the corresponding Github repositories. Statistics are averaged over five runs on each dataset. For VINS-MONO and S-MSCKF, the frontend and backend are run as separate ROS nodes. The lighter color represents the CPU usage of the frontend while the darker color represents the backend. Note that the backend of VINS-MONO is run at $10$hz because of limited CPU power.}
\label{fig: euroc benchmark}
\end{figure}
We compare our results on the EuRoC dataset with three representative VIO systems, OKVIS (stereo-optimization), ROVIO (monocular-filter), and VINS-MONO (monocular-optimization). Including the proposed method, the four visual inertial solutions are different combinations of monocular, stereo, filter-based, and optimization-based methods, which may provide insights into the pros and cons of the various approaches. For the monocular approaches, only the images from the left camera are used.  

Figure~\ref{fig: euroc benchmark} shows the Root Mean Square Error (RMSE) and the average CPU load of the four methods on the Dataset. The CPU load is measured with NUC6i7KYK equipped with quad-core i76770HQ 2.6Hz. The proposed method does not work properly on \verb!V2_03_difficult!. The reason is that we use the KLT optical flow algorithm~\cite{lucas1981iterative} for both temporal feature tracking and stereo matching to improve efficiency. The continuous inconsistency in brightness between the stereo images in \verb!V2_03_difficult! causes failures in the stereo feature matching, which then results in divergence of the filter. On the remaining datasets, the accuracy of the four different approaches is similar except ROVIO has larger error in the machine hall datasets which may be caused by the larger scene depth compared to the Vicon room datasets. 
For the CPU usage, the filter-based methods, both monocular and stereo, are more efficient compared with optimization based methods, which makes the filter-based approaches favorable for on-board real-time application. Between OKVIS and VINS-MONO, OKVIS has more CPU usage mainly because it uses Harris corner detector~\cite{harris1988combined} and BRISK~\cite{leutenegger2011brisk} descriptor for both temporal and stereo matching. Also, the backend of OKVIS is run at the fastest possible rate comparing to $10$Hz fixed in VINS-MONO. In the proposed S-MSCKF, around $80\%$ of the computation is caused by the frontend including feature detection, tracking and matching. The filter itself takes about $10\%$ of one core at $20$hz. Our proposed method provides a good compromise between accuracy and computational efficiency. 

%% file: tex/FastFlightDataset.tex
\subsection{Fast Flight Dataset}
\label{subsec: fast flight dataset}
\begin{figure}[t]
\centering
\begin{subfigure}[b]{0.23\textwidth}
\includegraphics[width=\textwidth]{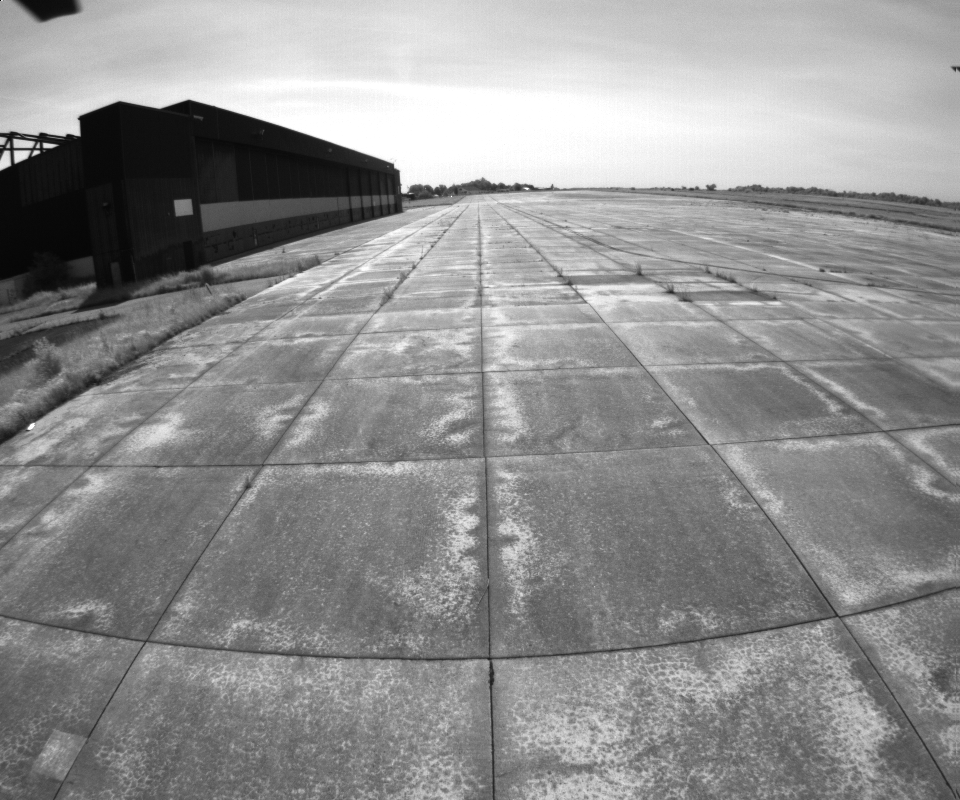}
\caption{•}
\label{fig: fast flight example img runway}
\end{subfigure}
\begin{subfigure}[b]{0.23\textwidth}
\includegraphics[width=\textwidth]{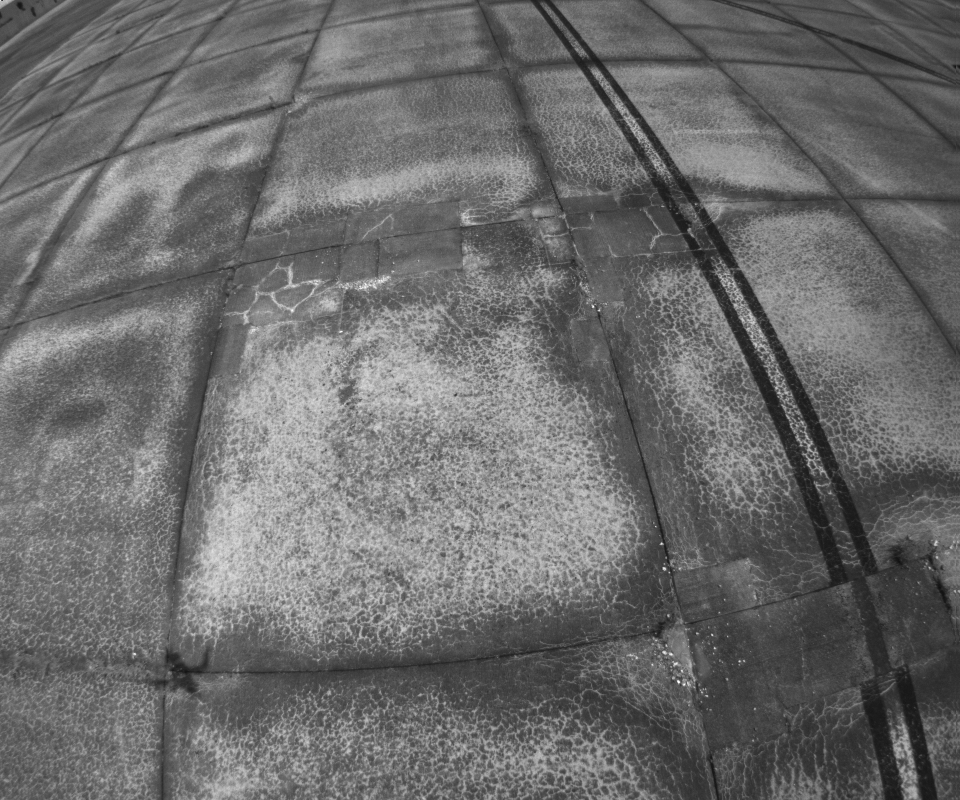}
\caption{•}
\label{fig: fast flight example img acceleration}
\end{subfigure}
\caption{Example images in the fast flight datasets. (a) images when the quadrotor is hovering. (b) images when the quadrotor is accelerating.}
\label{fig: fast flight example img}
\end{figure}

To further test the robustness of the proposed S-MSCKF, the algorithm is evaluated on four fast flight datasets with top speeds of $5$m/s, $10$m/s, $15$m/s, and $17.5$m/s respectively collected over an airport runway. During each run, the quadrotor is commanded to go to a waypoint $300$m ahead and return to the starting point. Our configuration includes two forward-looking PointGrey CM3-U3-13Y3M-CS cameras\footnote{https://www.ptgrey.com} running at $40$Hz with resolution $960\times 800$ and one VectorNav VN-100 Rugged IMU\footnote{http://www.vectornav.com/products/vn-100} running at $200$Hz. The whole sensor suite is synchronized based on the trigger signal from the IMU. To achieve proper image exposure under varying lighting conditions, the camera's internal auto-exposure is disabled and replaced by a fast-adapting external controller that maintains constant average image brightness. The controller uses only the left image for brightness measurement, but then applies identical shutter times and gains to both cameras simultaneously. Figure~\ref{fig: fast flight example img} shows some example images of the datasets. The dataset is publicly available at \url{https://github.com/KumarRobotics/msckf_vio/wiki}.

\begin{figure}[htp]
\centering
\begin{subfigure}[b]{0.4\textwidth}
\includegraphics[width=\textwidth]{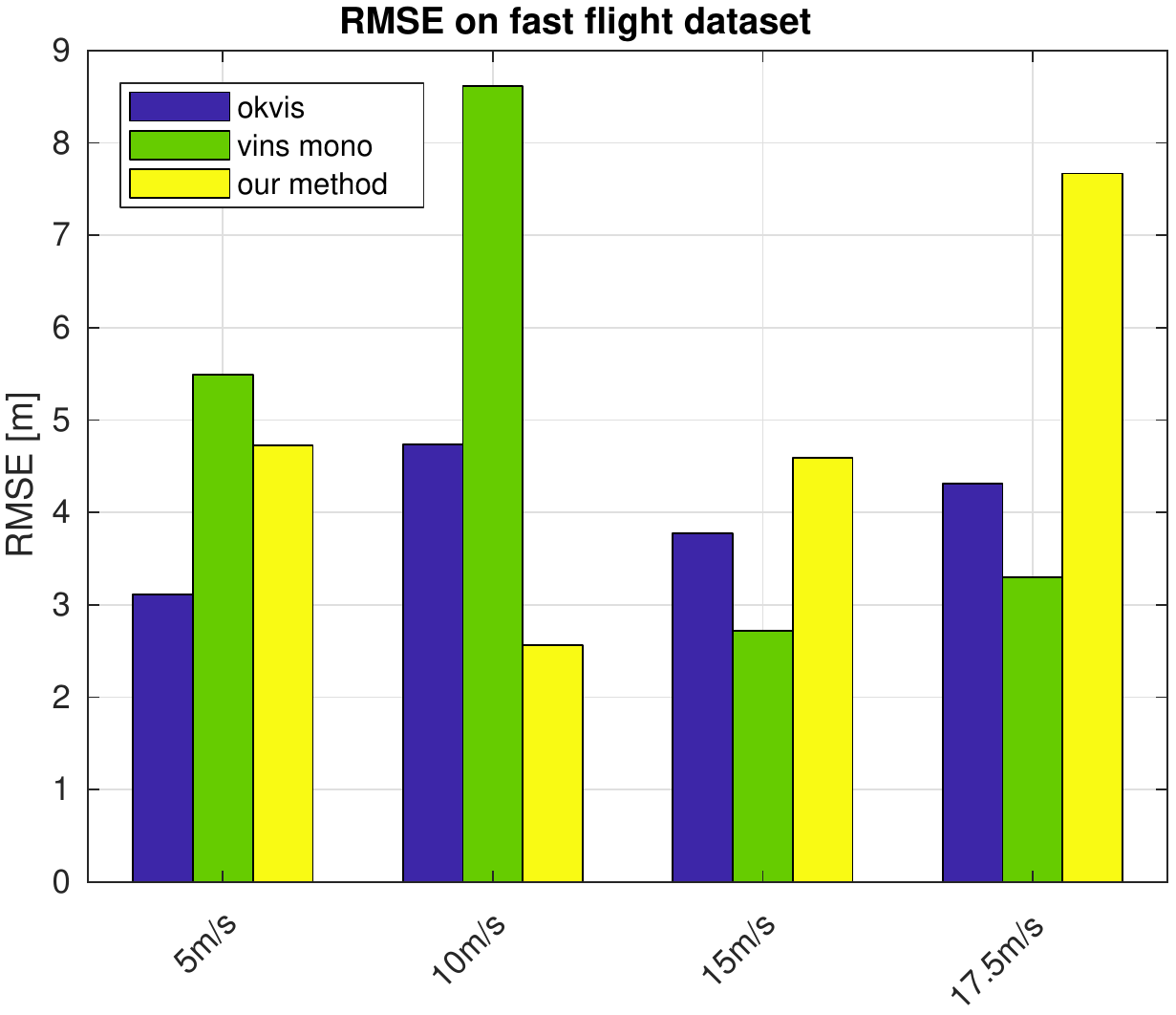}
\caption{•}
\label{fig: fast flight accuracy benchmark}
\end{subfigure}
\begin{subfigure}[b]{0.4\textwidth}
\includegraphics[width=\textwidth]{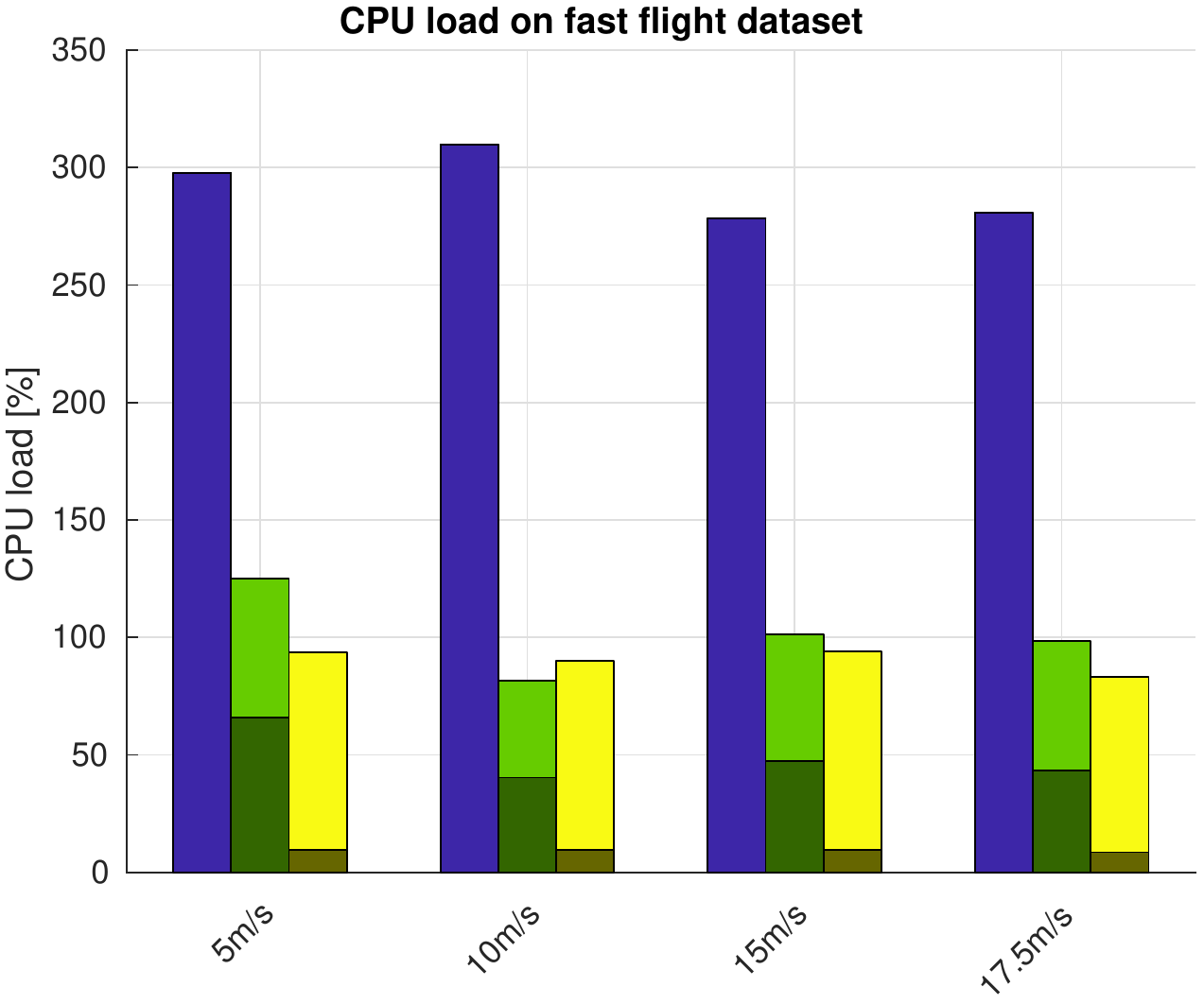}
\caption{•}
\label{fig: fast flight cpu benchmark}
\end{subfigure}
\caption{(a) RMSE and (b) CPU load of OKVIS, VINS-MONO, and the proposed method averaged over five runs on each dataset. As in the EuRoC dataset test, the CPU load of VINS-MONO and our method is shown as combinations of front and back end. The backend of VINS-MONO is run at $10$hz.}
\label{fig: fast flight benchmark}
\end{figure}

Figure~\ref{fig: fast flight benchmark} compares the accuracy and CPU usage of different VIO solutions on the fast flight dataset. The result of ROVIO is omitted in the comparison since it has significant drift in scale which results in much lower accuracy compared to other methods. The accuracy is evaluated by computing the RMSE of estimated and GPS position only in the $x$ and $y$ directions after proper alignment in both time and yaw. From the experiments, it can be observed that the S-MSCKF achieves the lowest CPU usage while maintaining similar accuracy comparing with other solutions. 

\begin{figure}[htp]
\centering
\begin{subfigure}[b]{0.22\textwidth}
\includegraphics[width=\textwidth]{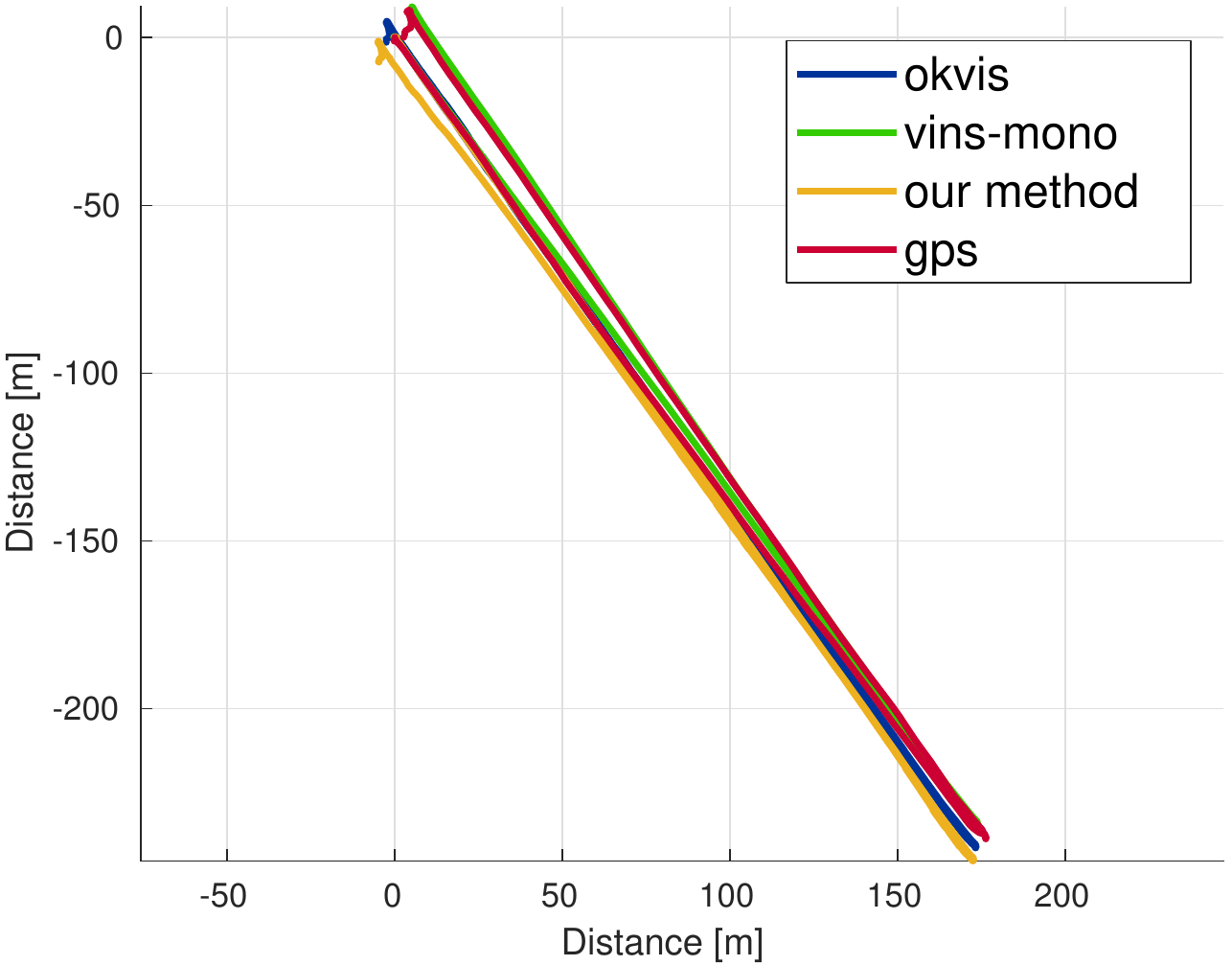}
\caption{}
\label{fig: fast flight trajectory}
\end{subfigure}
\begin{subfigure}[b]{0.22\textwidth}
\includegraphics[width=\textwidth]{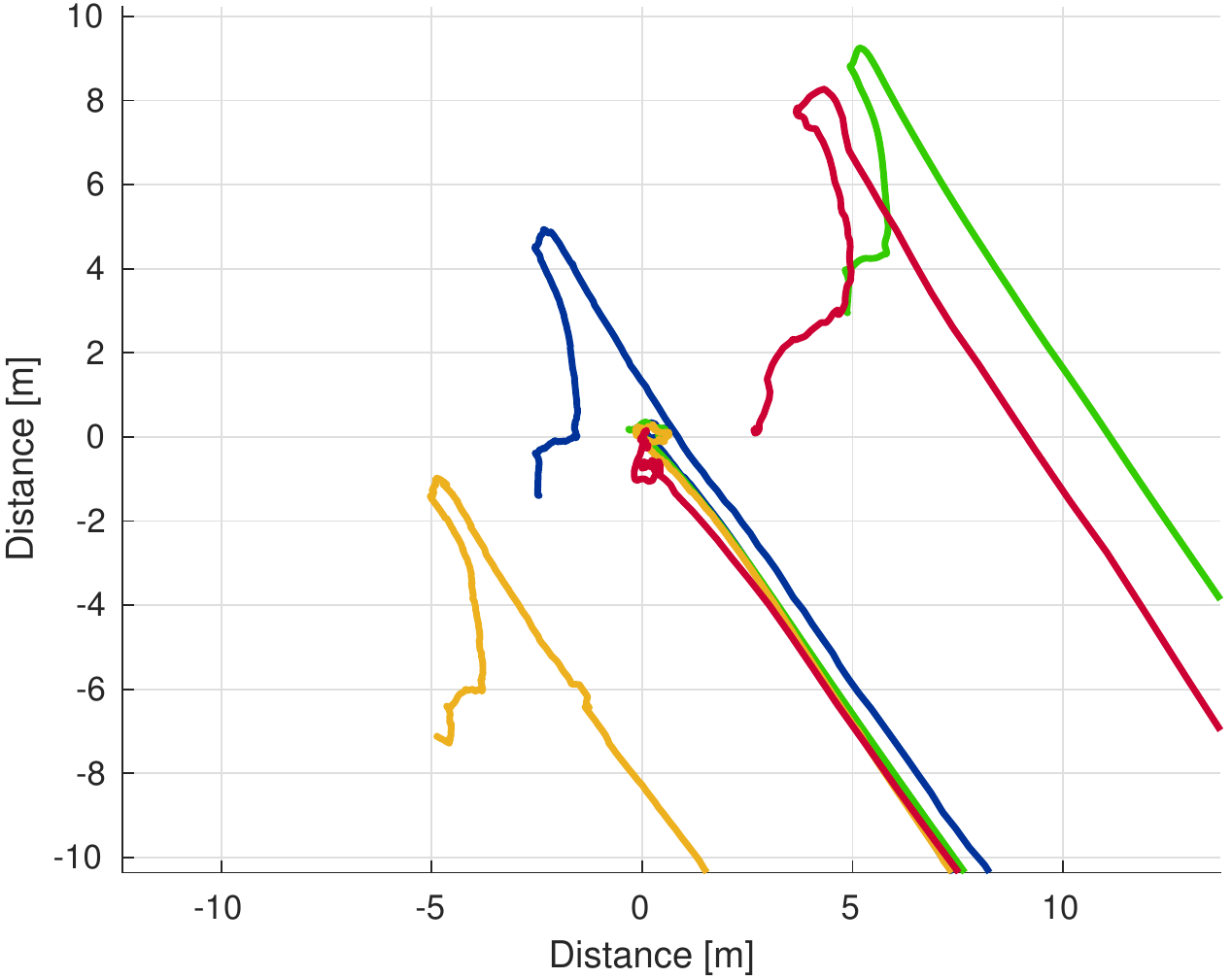}
\caption{}
\label{fig: fast flight trajectory origin}
\end{subfigure}\\
\begin{subfigure}[b]{0.22\textwidth}
\includegraphics[width=\textwidth]{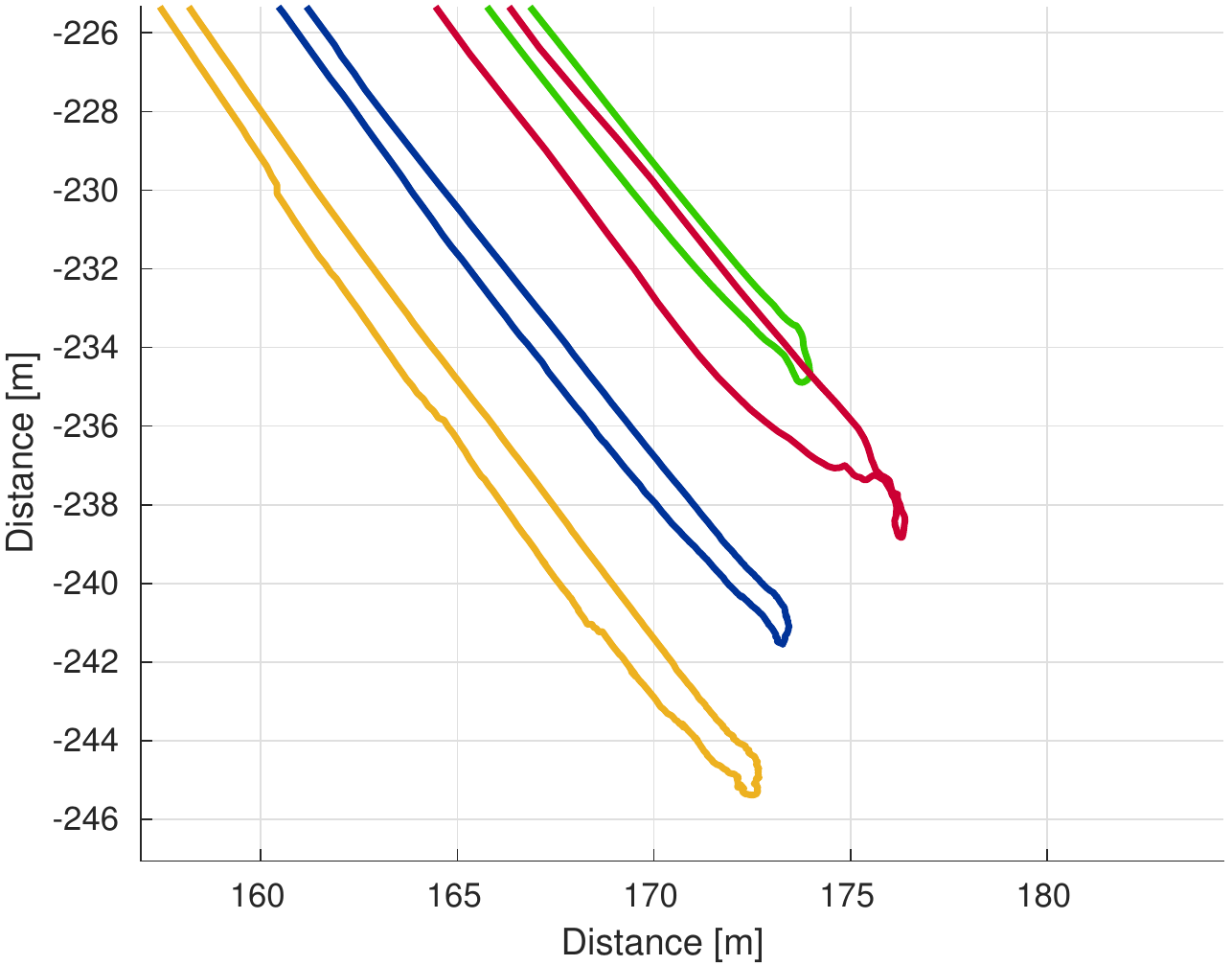}
\caption{}
\label{fig: fast flight trajectory goal}
\end{subfigure}
\begin{subfigure}[b]{0.22\textwidth}
\includegraphics[width=\textwidth]{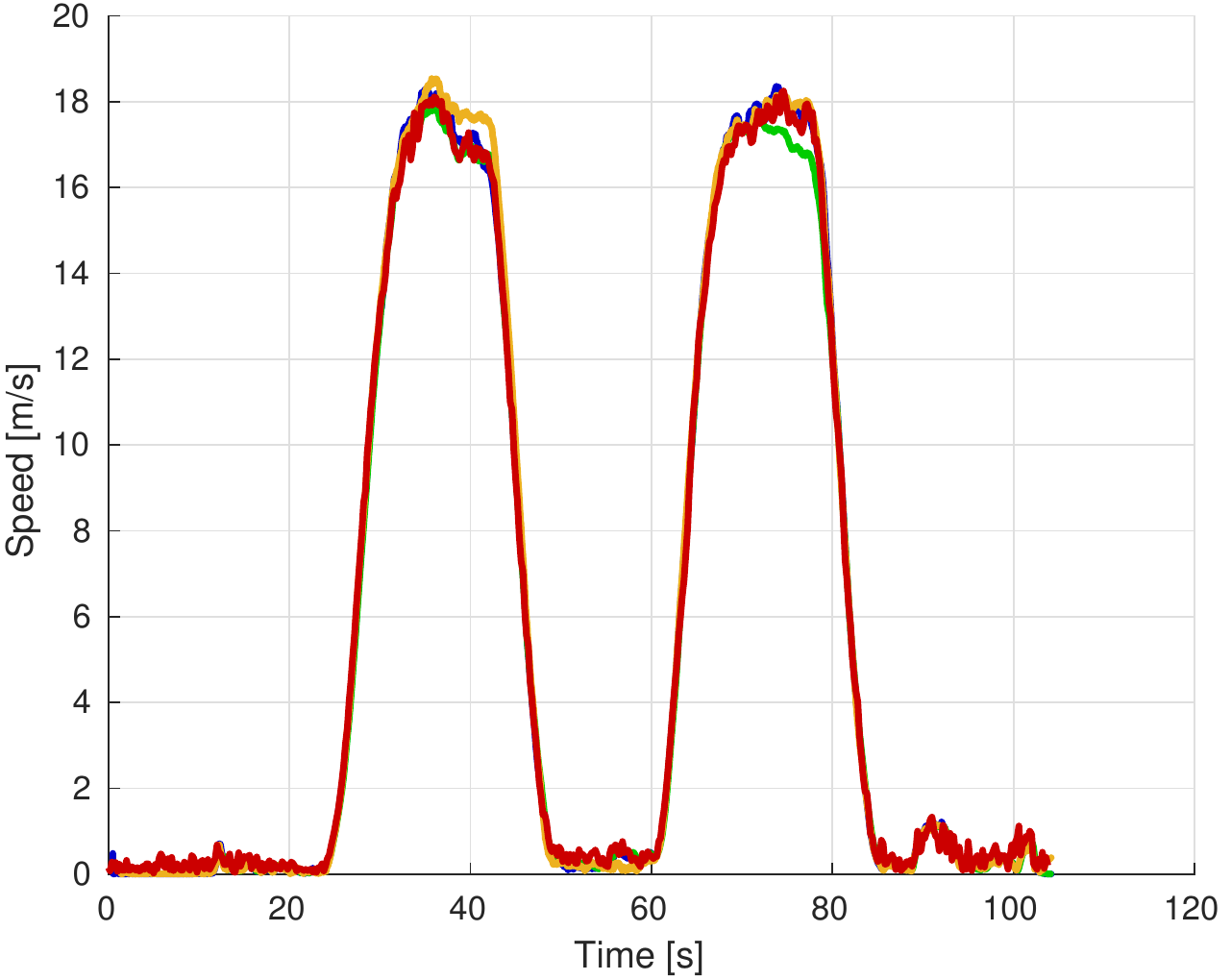}
\caption{}
\label{fig: fast flight max speed}
\end{subfigure}
\caption{(a) Aligned trajectories, (b) the starting point, (c) the goal location, and (d) speed profiles in the dataset with top speed at $17.5$m/s. }
\label{fig: max speed dataset}
\end{figure}

Note that compared to the experiments with the EuRoC dataset, the proposed method spends more computational effort on the image processing frontend.  One cause is the higher image frequency and resolution, while the other is that the aggressive flight induces shorter feature lifetime which then requires more frequent new feature detection. Figure~\ref{fig: max speed dataset} shows the aligned trajectories and speed profiles in the dataset with top speed at $17.5$m/s.

%% file: tex/FlaFieldTest.tex
\subsection{Autonomous Flight in Unstructured Environments}
\label{subsec: fla field test}
The proposed S-MSCKF has been thoroughly tested in various field experiments. In this section, we show an example of a fully autonomous flight where the robot has to first navigate through a wooded area, then look for an entrance into a warehouse, find a target, then return to the starting point. This experiment is illustrative since it includes a combination of different kinds of environments as well as common challenges for vision-based estimation including feature poverty, aggressive maneuvers, and significant changes in lighting conditions during indoor-outdoor transitions. 

\begin{figure}[htp]
\centering
\includegraphics[width=0.45\textwidth]{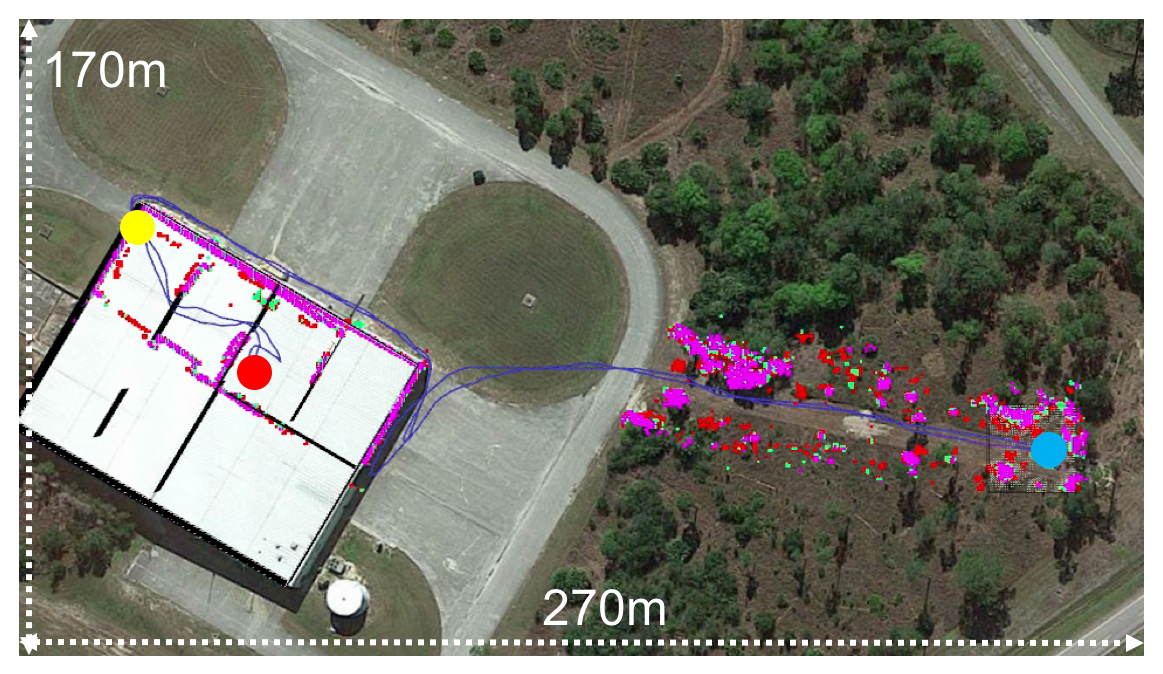}
\caption{The global map and round-trip trajectory overlaid on the Google satellite map in an fully autonomous flight experiment. The \textcolor{blue}{blue}, \textcolor{red}{red}, and \textcolor{yellow}{yellow} dots represents the staring point, goal location, and the only entrance of the warehouse respectively. The global laser point cloud is registered using the estimation produced by the S-MSCKF. Over $700$m trajectory, the final drift is around $3$m under $0.5\%$ of the total traveled distance.}
\label{fig: fla experiment iv}
\end{figure}

Figure \ref{fig: fla experiment iv} shows the global laser point cloud and round-trip trajectory overlaid on the Google satellite map.  Note that, during the experiment, the laser measurement is used for mapping only. The state estimation is solely based on the stereo cameras and IMU as the sensor configuration given in Section~\ref{subsec: fast flight dataset}. Over $700$m round-trip trajectory, the final drift is around $3$m, which is less than $0.5\%$ of the total traveled distance despite the combination of various challenges along the flight. More details of this trial can be found in the supplementary video\footnote{https://youtu.be/jxfJFgzmNSw}

%% file: tex/Conclusion.tex
\section{Conclusion}
\label{sec: conclusion}
In this paper, we present a new filter-based stereo visual inertial state estimation algorithm that uses the Multi-State Constraint Kalman Filter. We demonstrate the accuracy, efficiency, and robustness of our Stereo Multi-State Constraint Kalman Filter (S-MSCKF) using the EuRoC dataset as well as autonomous flight experiments in indoor and outdoor environments, comparing the computational efficiency and performance with state-of-art methods. We show that the S-MSCKF achieves robustness with a modest computational budget for aggressive, three-dimensional maneuvering, fast flight of speeds up to  $17.5$m/s, indoor/outdoor transition, and indoor navigation in cluttered environments. Finally, we describe the S-MSCKF implementation which is available at \url{https://github.com/KumarRobotics/msckf_vio}.

Since the global position and yaw is not observable in a VIO system as explained in Section~\ref{subsec: observability constraint}, the uncertainty of the corresponding directions will keep growing as the robot travels. It can be observed in the experiments that the filter estimation may jump or even diverge once the prior uncertainty is large. Our future work is addressing the possibility of planning intelligent trajectories for a quadrotor such that the growth in uncertainty of the VIO estimator is slower, which then helps extend the effective range of the robot.

%% file: tex/Appendices.tex
\section{}
\label{sec: error state dynamics}
The $\mathbf{F}$ and $\mathbf{G}$ in Eq.~\eqref{eq: error state dynamics} are,
\begin{equation*}
\mathbf{F} = 
\begin{pmatrix}
-\lfloor\hat{\bm{\omega}}{}_{\times}\rfloor & -\mathbf{I}_3 & 
\mathbf{0}_{3\times 3} & \mathbf{0}_{3\times 3} & \mathbf{0}_{3\times 3} \\
\mathbf{0}_{3\times 3} & \mathbf{0}_{3\times 3} & \mathbf{0}_{3\times 3} & 
\mathbf{0}_{3\times 3} & \mathbf{0}_{3\times 3} \\
-C\left({}^I_G\hat{\mathbf{q}}\right)^\top\lfloor\hat{\mathbf{a}}{}_{\times}\rfloor & 
\mathbf{0}_{3\times 3} & \mathbf{0}_{3\times 3} & 
-C\left({}^I_G\hat{\mathbf{q}}\right)^\top & \mathbf{0}_{3\times 3} \\
\mathbf{0}_{3\times 3} & \mathbf{0}_{3\times 3} & \mathbf{0}_{3\times 3} & 
\mathbf{0}_{3\times 3} & \mathbf{0}_{3\times 3} \\
\mathbf{0}_{3\times 3} & \mathbf{0}_{3\times 3} & \mathbf{I}_3 & 
\mathbf{0}_{3\times 3} & \mathbf{0}_{3\times 3} \\
\mathbf{0}_{3\times 3} & \mathbf{0}_{3\times 3} & \mathbf{0}_{3\times 3} & 
\mathbf{0}_{3\times 3} & \mathbf{0}_{3\times 3} \\
\mathbf{0}_{3\times 3} & \mathbf{0}_{3\times 3} & \mathbf{0}_{3\times 3} & 
\mathbf{0}_{3\times 3} & \mathbf{0}_{3\times 3}
\end{pmatrix}
\end{equation*}
and, 
\begin{equation*}
\mathbf{G} = 
\begin{pmatrix}
-\mathbf{I}_3 & \mathbf{0}_{3\times 3} & 
\mathbf{0}_{3\times 3} & \mathbf{0}_{3\times 3} \\
\mathbf{0}_{3\times 3} & \mathbf{I}_3 & 
\mathbf{0}_{3\times 3} & \mathbf{0}_{3\times 3} \\
\mathbf{0}_{3\times 3} & \mathbf{0}_{3\times 3} & 
-C\left({}^I_G\hat{\mathbf{q}}\right)^\top & \mathbf{0}_{3\times 3} \\
\mathbf{0}_{3\times 3} & \mathbf{0}_{3\times 3} & 
\mathbf{0}_{3\times 3} & \mathbf{0}_{3\times 3} \\
\mathbf{0}_{3\times 3} & \mathbf{0}_{3\times 3} & 
\mathbf{0}_{3\times 3} & \mathbf{I}_3 \\
\mathbf{0}_{3\times 3} & \mathbf{0}_{3\times 3} & 
\mathbf{0}_{3\times 3} & \mathbf{0}_{3\times 3} \\
\mathbf{0}_{3\times 3} & \mathbf{0}_{3\times 3} & 
\mathbf{0}_{3\times 3} & \mathbf{0}_{3\times 3}
\end{pmatrix}
\end{equation*}

\section{}
\label{sec: state augmentation jacobian}
The state augmentation Jacobian, $\mathbf{J}$, given in Eq.~\eqref{eq: state covariance augmentation}, is of the form,
\begin{equation*}
\mathbf{J} = 
\begin{pmatrix}
\mathbf{J}_I & \mathbf{0}_{6\times 6N}
\end{pmatrix}
\end{equation*}
where $\mathbf{J}_I$ is,
\begin{equation*}
\mathbf{J}_I = 
\begin{pmatrix}
C\left({}^I_G\hat{\mathbf{q}}\right) & \mathbf{0}_{3\times 9} & 
\mathbf{0}_{3\times 3} & \mathbf{I}_3 & \mathbf{0}_{3\times 3} \\
-C\left({}^I_G\hat{\mathbf{q}}\right)^\top \lfloor{}^I\hat{\mathbf{p}}_C {}_{\times}\rfloor & 
\mathbf{0}_{3\times 9} & \mathbf{I}_3 & \mathbf{0}_{3\times 3} & 
\mathbf{I}_{3}
\end{pmatrix}
\end{equation*}
Note that $\mathbf{J}_I$ given above corrects the typo in Eq. (16) of~\cite{mourikis2007multi}. 

\section{}
\label{sec: measurement jacobian}
Following the chain rule, $\mathbf{H}_{C_i}^j$ and $\mathbf{H}_{f_i}^j$, in Eq.~\eqref{eq: error measurement model}, can be computed as,
\begin{equation}
\label{eq: measurement jacobian}
\begin{gathered}
\mathbf{H}_{C_i}^j = 
\frac{\partial \mathbf{z}_i^j}{\partial {}^{C_{i,1}}\mathbf{p}_j} \cdot 
\frac{\partial {}^{C_{i,1}}\mathbf{p}_j}{\partial \mathbf{x}_{C_{i,1}}} + 
\frac{\partial \mathbf{z}_i^j}{\partial {}^{C_{i,2}}\mathbf{p}_j} \cdot 
\frac{\partial {}^{C_{i,2}}\mathbf{p}_j}{\partial \mathbf{x}_{C_{i,1}}} \\
\mathbf{H}_{f_i}^j = 
\frac{\partial \mathbf{z}_i^j}{\partial {}^{C_{i,1}}\mathbf{p}_j} \cdot 
\frac{\partial {}^{C_{i,1}}\mathbf{p}_j}{\partial {}^G\mathbf{p}_j} +
\frac{\partial \mathbf{z}_i^j}{\partial {}^{C_{i,2}}\mathbf{p}_j} \cdot 
\frac{\partial {}^{C_{i,2}}\mathbf{p}_j}{\partial {}^G\mathbf{p}_j} 
\end{gathered}
\end{equation}
where,
\begin{equation}
\label{eq: measurment jacobian expression}
\begin{gathered}
\frac{\partial \mathbf{z}_i^j}{\partial {}^{C_{i,1}}\mathbf{p}_j} = 
\frac{1}{{}^{C_{i, 1}}\hat{Z}_j}
\begin{pmatrix}
1 & 0 & -\frac{{}^{C_{i, 1}}\hat{X}_j}{{}^{C_{i, 1}}\hat{Z}_j} \\
0 & 1 & -\frac{{}^{C_{i, 1}}\hat{Y}_j}{{}^{C_{i, 1}}\hat{Z}_j} \\
0 & 0 & 0 \\
0 & 0 & 0 
\end{pmatrix} \\
\frac{\partial \mathbf{z}_i^j}{\partial {}^{C_{i,2}}\mathbf{p}_j} = 
\frac{1}{{}^{C_{i, 2}}\hat{Z}_j}
\begin{pmatrix}
0 & 0 & 0 \\
0 & 0 & 0 \\
1 & 0 & -\frac{{}^{C_{i, 2}}\hat{X}_j}{{}^{C_{i, 1}}\hat{Z}_j} \\
0 & 1 & -\frac{{}^{C_{i, 2}}\hat{Y}_j}{{}^{C_{i, 1}}\hat{Z}_j} 
\end{pmatrix} \\
\frac{\partial {}^{C_{i,1}}\mathbf{p}_j}{\partial \mathbf{x}_{C_{i,1}}} = 
\begin{pmatrix}
\lfloor{}^{C_{i,1}}\hat{\mathbf{p}}_j{}_{\times}\rfloor & 
-C\left({}^{C_{i,1}}_G\hat{\mathbf{q}}\right)
\end{pmatrix} \\
\frac{\partial {}^{C_{i,1}}\mathbf{p}_j}{\partial {}^G\mathbf{p}_j} = 
C\left({}^{C_{i,1}}_G\hat{\mathbf{q}}\right) \\
\frac{\partial {}^{C_{i,2}}\mathbf{p}_j}{\partial \mathbf{x}_{C_{i,1}}} = 
C\left({}^{C_{i,1}}_{C_{i,2}}\mathbf{q}\right)^\top
\begin{pmatrix}
\lfloor{}^{C_{i,1}}\hat{\mathbf{p}}_j{}_{\times}\rfloor & 
-C\left({}^{C_{i,1}}_G\hat{\mathbf{q}}\right)
\end{pmatrix} \\
\frac{\partial {}^{C_{i,2}}\mathbf{p}_j}{\partial {}^G\mathbf{p}_j} = 
C\left({}^{C_{i,1}}_{C_{i,2}}\mathbf{q}\right)^\top
C\left({}^{C_{i,1}}_G\hat{\mathbf{q}}\right)
\end{gathered}
\end{equation}

\section{}
\label{sec: nullify measurement jacobian}
By defining the following short-hand notation from Eq.~\eqref{eq: measurment jacobian expression}
\begin{equation*}
\begin{gathered}
\frac{\partial \mathbf{z}_i^j}{\partial {}^{C_{i,1}}\mathbf{p}_j} = 
\begin{pmatrix}
\mathbf{J}_1 \\ \mathbf{0}
\end{pmatrix}, \quad
\frac{\partial \mathbf{z}_i^j}{\partial {}^{C_{i,2}}\mathbf{p}_j} = 
\begin{pmatrix}
\mathbf{0} \\ \mathbf{J}_2
\end{pmatrix}\\
\frac{\partial {}^{C_{i,1}}\mathbf{p}_j}{\partial \mathbf{x}_{C_{i,1}}} = 
\mathbf{H}_1, \quad 
\frac{\partial {}^{C_{i,1}}\mathbf{p}_j}{\partial {}^G\mathbf{p}_j} = 
\mathbf{H}_2, \quad
C\left({}^{C_{i,1}}_{C_{i,2}}\mathbf{q}\right) = 
\mathbf{R}\ ,
\end{gathered}
\end{equation*}
the measurement Jacobian in Eq.~\eqref{eq: measurement jacobian} can be compactly written as
\begin{equation*}
\mathbf{H}_{C_i}^j = 
\begin{pmatrix}
\mathbf{J}_1 \mathbf{H}_1 \\
\mathbf{J}_2 \mathbf{R}^\top \mathbf{H}_1
\end{pmatrix},\quad
\mathbf{H}_{f_i}^j =
\begin{pmatrix}
\mathbf{J}_1 \mathbf{H}_2 \\
\mathbf{J}_2 \mathbf{R}^\top \mathbf{H}_2
\end{pmatrix}\ .
\end{equation*}
Assuming $\mathbf{v} = \left(\mathbf{v}_1^\top,\ \mathbf{v}_2^\top\right)^\top\in\mathbb{R}^4$ is the left null space of $\mathbf{H}_{f_i}^j$, then,
\begin{equation*}
\mathbf{v}^\top \mathbf{H}_{f_i}^j  = 
\left(\mathbf{v}_1^\top \mathbf{J}_1 + 
\mathbf{v}_2^\top\mathbf{J}_2\mathbf{R}^\top\right) 
\mathbf{H}_2 = \mathbf{0}
\end{equation*}
Since $\mathbf{H}_2 = C\left({}^{C_{i,1}}_G\hat{\mathbf{q}}\right)$ is a rotation matrix, $\text{rank}\left(\mathbf{H}_2\right) = 3$ which implies that $\mathbf{v}_1^\top \mathbf{J}_1 + \mathbf{v}_2^\top\mathbf{J}_2\mathbf{R}^\top = \mathbf{0}$. With such property, it immediately follows that $\mathbf{v}$ is also the left null space of $\mathbf{H}_{C_i}^j$, 
\begin{equation*}
\mathbf{v}^\top \mathbf{H}_{C_i}^j = 
\left(\mathbf{v}_1^\top \mathbf{J}_1 + 
\mathbf{v}_2^\top\mathbf{J}_2\mathbf{R}^\top\right) 
\mathbf{H}_1 = \mathbf{0}
\end{equation*}
Therefore, a singe stereo measurement cannot be directly used for measurement update.